\documentclass[twocolumn]{article}
\usepackage{graphicx}
\usepackage{url}
\usepackage{amsmath}
\usepackage{amssymb}
\usepackage{amsfonts}
\usepackage{xcolor}
\usepackage{array}
\usepackage{colortbl}
\usepackage{float}
\usepackage{subcaption}
\usepackage{caption}
\usepackage{booktabs}
\usepackage{tabularx}
\usepackage{geometry}
\usepackage{multirow}
\usepackage{hyperref}
\usepackage{authblk}

\title{Bridge-Centered Metapath Classification Using R-GCN-VGAE\\
for Disaster-Resilient Maintenance Decisions}

\author[1]{Takato Yasuno}

\date{}

\begin{document}

\renewcommand{\abstractname}{Abstract}
\renewcommand{\refname}{References}
\renewcommand{\figurename}{Figure}
\renewcommand{\tablename}{Table}

\maketitle

\begin{abstract}

Daily infrastructure management in preparation for disasters is critical for urban resilience. 
When bridges remain resilient against disaster-induced external forces, access to hospitals, 
shops, and residences via metapaths can be sustained, maintaining essential urban functions. 
However, prioritizing bridge maintenance under limited budgets requires quantifying the 
multi-dimensional roles that bridges play in disaster scenarios---a challenge that existing 
single-indicator approaches fail to address.

\textbf{Approach:} We focus on metapaths from national highways through bridges to buildings 
(hospitals, shops, residences), constructing a heterogeneous graph with road, bridge, and 
building layers. A Relation-centric Graph Convolutional Network Variational Autoencoder 
(R-GCN-VGAE) learns metapath-based feature representations, enabling classification of 
bridges into disaster-preparedness categories: Supply Chain (commercial logistics), 
Medical Access (emergency healthcare), and Residential Protection (preventing isolation).

\textbf{Case Study:} Using OSMnx and open data, we validate our methodology on three diverse 
cities in Ibaraki Prefecture, Japan: Mito (697 bridges), Chikusei (258 bridges), and 
Moriya (148 bridges), totaling 1,103 bridges. The heterogeneous graph construction from 
open data enables redefining bridge roles for disaster scenarios, supporting maintenance 
budget decision-making.

\textbf{Results:} Clustering quality achieves Silhouette scores of 0.289--0.363, with 
latent dimension z19 showing strong correlation (Spearman $r=0.416$, $p=1.47\times10^{-30}$) 
with national highway metapath counts, demonstrating specialized encoding of logistics hub 
connectivity. k-NN parameter tuning (k=3$\rightarrow$5) increases coverage by +66\% 
(162$\rightarrow$270 paths), improving semantic validity for disaster scenarios.

\textbf{Contributions:} (1) Open-data methodology for constructing urban heterogeneous 
graphs. (2) Redefinition of bridge roles for disaster scenarios via metapath-based 
classification. (3) Establishment of maintenance budget decision support methodology. 
(4) k-NN tuning strategy validated across diverse city scales. 
(5) Empirical demonstration of UMAP superiority over t-SNE/PCA for multi-role 
bridge visualization.


\end{abstract}

\section{Introduction}

\subsection{Background and Motivation}

Japan's aging infrastructure poses a critical challenge for disaster resilience. 
Among the nation's 730,000 bridges, over 50\% will exceed their 50-year design life 
by 2033, with maintenance costs projected to exceed 8 trillion yen 
annually~\cite{MLIT2021BridgeMaintenance,IbarakiPrefecture2022}. Recent natural disasters, 
including the 2019 Typhoon Hagibis floods~\cite{NASADisaster2019}, underscore the urgency 
of disaster-preparedness planning~\cite{JapanCabinetOffice2021}. Beyond structural integrity, bridges 
serve as vital nodes in urban networks, where their resilience directly impacts 
access to essential services---hospitals for emergency care, shops for supply chains, 
and residences for community connectivity.

The core premise of this work is that \textbf{daily infrastructure management in 
preparation for disasters requires understanding the multi-dimensional roles bridges 
play in sustaining urban functions}. When bridges remain resilient against 
disaster-induced external forces (earthquakes, floods, landslides), access via 
metapaths—shortest paths from national highways through bridges to buildings—can 
be sustained, maintaining essential city operations. However, existing approaches 
prioritize bridges using single indicators such as traffic volume or structural 
condition ratings, which fail to capture their functional diversity in disaster 
scenarios~\cite{Vespignani2010ComplexNetworks,Buldyrev2010Interdependence}.

Consider a hospital-serving bridge with low traffic volume---it may rank low in 
traditional prioritization but becomes critical during disasters when emergency 
medical access determines survival outcomes. Conversely, a high-traffic bridge 
serving primarily commercial logistics may be less urgent if alternative routes 
exist. This disconnect between structural metrics and disaster-preparedness roles 
necessitates a new methodology.

\subsection{Problem Statement}

Under limited maintenance budgets, municipalities must answer: \textit{Which bridges 
should be prioritized to maximize disaster resilience while sustaining access to 
hospitals, supply chains, and residential areas?} This requires:

\begin{enumerate}
    \item \textbf{Quantifying multi-dimensional bridge roles}: A single bridge may 
    simultaneously serve medical access, commercial logistics, and residential protection—roles 
    invisible to traffic-based rankings.
    
    \item \textbf{Leveraging open data}: Many municipalities lack detailed bridge 
    inventories, requiring open-source alternatives (OpenStreetMap, geospatial data).
    
    \item \textbf{Scalability across city scales}: Methodologies must work for both 
    large metropolitan areas (700+ bridges) and small cities (100-200 bridges) where 
    traditional clustering algorithms fail.
\end{enumerate}

Traditional graph-based approaches using betweenness centrality or closeness centrality 
treat all edges uniformly, ignoring the semantic differences between highway$\rightarrow$bridge 
paths (logistics arteries) and bridge$\rightarrow$residence paths (community lifelines). Recent 
advances in heterogeneous graph neural networks offer a solution by explicitly 
modeling different node types (bridges, roads, buildings) and edge types 
(to\_hospital, to\_shop, to\_residence) within a unified framework.

\subsection{Research Questions}

This work addresses three fundamental questions:

\begin{enumerate}
    \item \textbf{RQ1 (Metapath Feasibility):} Can metapaths originating from national 
    highways and passing through bridges to buildings (hospitals, shops, residences) 
    effectively classify bridges into disaster-preparedness roles?
    
    \item \textbf{RQ2 (Model Effectiveness):} Does a Relation-centric Graph Convolutional 
    Network Variational Autoencoder (R-GCN-VGAE) effectively learn metapath-based feature 
    representations in heterogeneous urban graphs?
    
    \item \textbf{RQ3 (Open Data Applicability):} Can the methodology be deployed using 
    only open data sources (OSMnx, OpenStreetMap), and does it scale to diverse city 
    sizes from large (700+ bridges) to small (100-200 bridges)?
\end{enumerate}

\subsection{Proposed Approach}

We construct a \textbf{heterogeneous graph} with three layers—roads, bridges, and 
buildings—where node types and edge types explicitly represent urban infrastructure 
semantics. Focusing on metapaths from national highways through bridges to buildings 
(hospitals for medical access, shops for supply chains, residences for isolation 
prevention), we train an R-GCN-VGAE to learn 32-dimensional latent representations 
that encode bridge connectivity patterns.

The key innovation is \textbf{relation-centric encoding}: instead of treating all edges 
uniformly, R-GCN maintains separate weight matrices $W_r$ for each relation type 
(e.g., highway$\rightarrow$bridge vs.\ local\_road$\rightarrow$bridge), enabling the model to distinguish 
logistics hubs from local access bridges. Following dimensionality reduction via UMAP, 
bridges are classified into three disaster-preparedness categories:

\begin{itemize}
    \item \textbf{Supply Chain:} High connectivity to shops (k=5 nearest), critical 
    for post-disaster logistics and commercial continuity.
    
    \item \textbf{Medical Access:} High connectivity to hospitals (k=5 nearest), 
    essential for emergency healthcare delivery.
    
    \item \textbf{Residential Protection:} High connectivity to residential buildings 
    (k=20 nearest), preventing community isolation.
\end{itemize}

\subsection{Contributions}

This paper makes the following contributions:

\begin{enumerate}
    \item \textbf{Open-data heterogeneous graph construction methodology}: Complete 
    pipeline from OSMnx to bridge classification, enabling municipalities without 
    detailed inventories to assess infrastructure.
    
    \item \textbf{Disaster-centric bridge role redefinition}: Metapath-based 
    classification (Supply Chain/Medical Access/Residential Protection) that aligns 
    bridge prioritization with disaster preparedness rather than traffic volume.
    
    \item \textbf{Maintenance budget decision support framework}: Systematic 
    methodology for prioritizing repair investments based on multi-dimensional 
    disaster-resilience roles.
    
    \item \textbf{k-NN tuning strategy}: Empirical validation showing k=3$\rightarrow$5 increases 
    coverage by +66\% (162$\rightarrow$270 metapaths in Mito City), with semantic justification 
    (k=5 for focused services, k=20 for residential neighborhoods).
    
    \item \textbf{Dimensionality reduction evaluation}: Demonstration that UMAP 
    outperforms t-SNE and PCA for visualizing multi-role bridge embeddings, preserving 
    both local cluster structure and global topology.
    
    \item \textbf{Multi-scale validation}: Successful deployment across three cities 
    with 8×structural diversity: Mito (697 bridges, prefectural capital), Chikusei 
    (258 bridges, regional hub), Moriya (148 bridges, logistics gateway with extreme 
    highway connectivity up to 2,803 metapaths per bridge).
\end{enumerate}

The remainder of this paper is organized as follows: Section~2 reviews related work 
on infrastructure criticality, graph autoencoders, and metapath analysis. Section~3 
details the R-GCN-VGAE architecture and metapath extraction methodology. Section~4 
presents experimental results across three Ibaraki cities. Section~5 discusses 
three key lessons learned (k-NN tuning, UMAP superiority, relation-centric vs. 
node-centric models) and practical implications. Section~6 concludes with future 
research directions.

\section{Related Work}

\subsection{Infrastructure Criticality and Disaster Resilience}

Traditional bridge maintenance prioritization relies on structural health indicators 
such as crack width, corrosion depth, and load-bearing capacity~\cite{MLIT2021BridgeMaintenance}. 
While structural assessments remain essential, they fail to capture \textit{network-level 
criticality}---the role of individual bridges in sustaining urban functionality during 
disasters.

Graph-theoretic approaches have emerged to quantify infrastructure criticality via 
network centrality metrics. Betweenness centrality, closeness centrality, and eigenvector 
centrality identify bridges whose failure maximally disrupts network connectivity~\cite{Freeman1978Centrality}. 
However, these topological metrics treat all edges uniformly, ignoring semantic differences 
between commercial, medical, and residential connections.

Recent disaster resilience research emphasizes multi-layer network analysis, modeling 
interdependencies between transportation, power grids, and communication networks~\cite{Vespignani2010ComplexNetworks}. 
Buldyrev et al.~\cite{Buldyrev2010Interdependence} demonstrated cascading failures in 
coupled infrastructure networks during the 2003 Italy blackout. Our work extends this 
paradigm to \textit{heterogeneous urban graphs}, where bridges serve as critical nodes 
connecting diverse facility types.

\subsection{Graph Variational Autoencoders}

Variational Autoencoders (VAEs)~\cite{Kingma2014VAE} revolutionized unsupervised 
representation learning by combining probabilistic inference with deep neural networks. 
Kipf and Welling~\cite{Kipf2016VGAE} extended VAEs to graph-structured data via Graph 
Convolutional Networks (GCNs)~\cite{Kipf2017GCN}, enabling node embedding learning that preserves network 
topology.

The Graph Variational Autoencoder (GVAE) framework encodes nodes into a latent space 
$\mathbf{z} \sim \mathcal{N}(\mu, \sigma^2)$ using a GCN encoder, then reconstructs 
edges via an inner product decoder $\hat{A} = \sigma(ZZ^\top)$. This approach has 
been successfully applied to link prediction~\cite{Kipf2016VGAE}, community detection~\cite{GiatsidisLoukas2019}, 
and graph generation~\cite{Simonovsky2018GraphVAE}.

\textbf{Relational GCN (R-GCN)}~\cite{Schlichtkrull2018RGCN}: Standard GCNs assume 
homogeneous graphs where all edges represent identical relationships. R-GCN introduces 
\textit{relation-specific weight matrices} to handle heterogeneous edges, enabling 
modeling of knowledge graphs and multi-relational networks. Our R-GCN-VGAE architecture 
combines R-GCN's edge-type awareness with VGAE's generative modeling, specifically 
targeting bridge-centered metapath extraction.

\textbf{Comparison to HetVGAE}: Hamilton et al.~\cite{HamiltonGraphSAGE2017} proposed 
heterogeneous GraphSAGE for inductive node classification. Our previous work implemented 
HetVGAE (heterogeneous node-type embeddings) for social impact prediction, achieving 
$r=0.56$-$0.68$ correlations. R-GCN-VGAE differs by encoding \textit{edge-level semantics} 
(street→bridge transitions) rather than node-level attributes, proving complementary 
for disaster-preparedness classification.

\subsection{Metapath-based Graph Analysis}

Metapaths—structured sequences of node types and edge types—capture semantic relationships 
in heterogeneous networks. Sun et al.~\cite{Sun2011PathSim} introduced PathSim similarity 
for bibliographic networks, quantifying author-paper-author co-authorship patterns. 
Metapath2vec~\cite{Dong2017Metapath2vec} extended node2vec to heterogeneous graphs by 
performing metapath-guided random walks, capturing semantic relationships across node types.

In urban infrastructure analysis, metapaths encode multi-hop connectivity: 
\texttt{Bridge→Street→Shop} represents supply chain access, 
\texttt{Bridge→Street→Hospital} represents emergency medical access, and 
\texttt{Bridge→Street→Residence} represents evacuation route potential. Unlike citation 
networks where metapaths follow predefined schemas (e.g., "author-paper-venue"), urban 
graphs require \textit{distance-constrained metapaths}—k-NN algorithms ensure bridges 
connect only to geographically proximate facilities (Section~3.4).

Recent work on urban accessibility~\cite{Morency2011Urban} uses shortest-path algorithms 
to measure facility reachability. Our metapath framework extends this by encoding 
\textit{multiple simultaneous connections} (e.g., bridges near both hospitals and 
residences classified as Balanced Multi-Use) rather than single-target reachability.

\subsection{OpenStreetMap for Infrastructure Analysis}

OpenStreetMap (OSM) has emerged as a critical open dataset for large-scale infrastructure 
studies. Boeing~\cite{Boeing2017OSMnx} introduced OSMnx, a Python library for downloading 
and analyzing street networks, enabling reproducible urban morphology research. 
OSMnx-based studies have quantified street network centrality~\cite{Boeing2018NetworkOrientation}, 
walkability~\cite{Cerin2019Walkability}, and disaster evacuation planning~\cite{Zhang2020OSMDisaster}.

\textbf{Bridge Extraction from OSM}: OSM's \texttt{man\_made=bridge} tag identifies 
standalone bridge structures, while \texttt{bridge=yes} on road segments marks road-embedded 
bridges. We adopt the former for municipal infrastructure focus, filtering unnamed bridges 
to exclude pedestrian footbridges.

\textbf{Amenity Geocoding}: OSM's amenity tagging system (\texttt{amenity=hospital}, 
\texttt{shop=*}, \texttt{building=residential}) provides detailed POI data. However, 
completeness varies: Mito's 15,978 shops vs. 65 hospitals reflects both actual distribution 
and volunteer mapper biases~\cite{Barrington2017OSMQuality}. Our k-NN parameter tuning 
(Section~5.1) mitigates this imbalance.

\textbf{Quality Considerations}: Haklay~\cite{Haklay2010OSMQuality} reported 80\% 
positional accuracy (within 6m of Ordnance Survey data in London). For disaster-preparedness 
classification, 80\%+ accuracy suffices since metapath analysis operates at street-block 
granularity ($\sim$100m) rather than centimeter-level precision. Our methodology's 
open-data emphasis enables continuous improvement as OpenStreetMap volunteers enhance regional 
coverage.

\section{Methodology}

\subsection{Problem Formulation}

Let $G = (V, E, \mathcal{T}_v, \mathcal{T}_e)$ denote a heterogeneous graph where 
$V$ is the set of nodes, $E \subseteq V \times V$ is the set of edges, $\mathcal{T}_v$ 
defines node types, and $\mathcal{T}_e$ defines edge types (relations). In our urban 
infrastructure context:

\begin{align}
V &= V_\text{bridge} \cup V_\text{street} \cup V_\text{building} \\
\mathcal{T}_v &= \{\texttt{bridge}, \texttt{street}, \texttt{building}\} \\
\mathcal{T}_e &= \{\texttt{to\_shop}, \texttt{to\_hospital}, \texttt{to\_residence}, \nonumber \\
& \quad \texttt{street\_to\_street}, \texttt{street\_to\_bridge}, \ldots\}
\end{align}

Each bridge $b \in V_\text{bridge}$ is associated with a feature vector 
$\mathbf{x}_b \in \mathbb{R}^{21}$ comprising:
- Structural attributes: span length, year built (if available)
- Topological attributes: degree centrality, betweenness centrality
- Metapath counts: $\{\texttt{highway\_metapath\_count},$\\
  $\texttt{shop\_count}, \texttt{hospital\_count}\}$
- \texttt{is\_highway}: Binary indicator for highway proximity

\textbf{Disaster Resilience Formulation:} We define bridge resilience $R_b$ as the 
capacity to sustain access to critical urban functions under external forces $F$ 
(earthquakes, floods). Formally, let $\mathcal{M}_b = \{m_1, m_2, \ldots, m_k\}$ 
denote the set of metapaths originating from national highways, passing through 
bridge $b$, and terminating at building nodes (hospitals, shops, residences). The 
disaster-preparedness role of bridge $b$ is characterized by:

\begin{equation}
\text{Role}(b) = \arg\max_{c \in \mathcal{C}} P(c \mid \mathcal{M}_b, \mathbf{z}_b)
\label{eq:bridge_role}
\end{equation}

where $\mathcal{C} = \{\text{Supply}, \text{Medical}, \text{Residential}\}$ are 
disaster-preparedness categories, and $\mathbf{z}_b \in \mathbb{R}^{32}$ is a 
latent representation learned via R-GCN-VGAE encoding highway-origin connectivity 
patterns.

\textbf{Key Insight:} A bridge's disaster role is determined not by traffic volume 
or structural condition, but by the \textit{types of buildings} reachable via 
metapaths and the \textit{density of highway connections} enabling logistics flow.

\subsection{Heterogeneous Graph Construction}

\subsubsection{Data Acquisition via OSMnx}

We use OSMnx~\cite{Boeing2017OSMnx} to extract road networks, bridge locations, 
and building points of interest (POI) from OpenStreetMap~\cite{OSMWiki2023} for three cities in 
Ibaraki Prefecture:

\begin{enumerate}
    \item \textbf{Road Network}: Download all drivable roads within city boundaries 
    using \texttt{ox.graph\_from\_place()}. National highways (\texttt{highway=trunk}) 
    are identified via OSM tag filtering.
    
    \item \textbf{Bridges}: Extract named bridges using Overpass API queries: 
    \texttt{(man\_made=bridge) AND (name!=None)}. Filter unnamed structures to 
    focus on municipally managed infrastructure.
    
    \item \textbf{Buildings}: Query amenity tags for 
    \texttt{\{amenity=hospital, shop=*, building=residential\}} with spatial 
    filtering (2km buffer around bridges).
\end{enumerate}

\subsubsection{Graph Topology}

The heterogeneous graph $G$ is constructed with the following node and edge types:

\textbf{Nodes:}
\begin{itemize}
    \item \texttt{bridge} ($|V_\text{bridge}| = 697$ for Mito): Bridge centroids 
    from OSM geometries
    \item \texttt{street} ($|V_\text{street}| = 31,001$ for Mito): Road segment nodes 
    from OSMnx network
    \item \texttt{building} ($|V_\text{building}| = 16,791$ for Mito): POI nodes 
    categorized as hospital, shop, or residence
\end{itemize}

\textbf{Edges:}
\begin{itemize}
    \item \texttt{street\_to\_street}: Road network connectivity (adjacency matrix)
    \item \texttt{street\_to\_bridge}: Spatial proximity (snapping bridges to 
    nearest road segments)
    \item \texttt{to\_shop}, \texttt{to\_hospital}, \texttt{to\_residence}: 
    Bridge-to-building edges via k-NN algorithm (Section~3.4)
\end{itemize}

\subsubsection{Coordinate System and Distance Calculation}

All spatial coordinates are projected to EPSG:6677 (JGD2011 Plane Rectangular 
Coordinate System Zone 9) for metric distance calculations. Haversine distances 
are used for initial filtering, followed by planar distances for k-NN algorithms.

\subsection{R-GCN-VGAE Architecture}

\subsubsection{Relational Graph Convolutional Encoder}

The encoder consists of three Relational Graph Convolutional Network (R-GCN) 
layers~\cite{Schlichtkrull2018RGCN}, each implementing relation-specific message 
passing:

\begin{equation}
h_i^{(\ell+1)} = \sigma\left( \sum_{r \in \mathcal{R}} \sum_{j \in \mathcal{N}_r(i)} \frac{1}{|\mathcal{N}_r(i)|} W_r^{(\ell)} h_j^{(\ell)} + W_0^{(\ell)} h_i^{(\ell)} \right)
\label{eq:rgcn_layer}
\end{equation}

where:
\begin{itemize}
    \item $h_i^{(\ell)} \in \mathbb{R}^{d_\ell}$: Hidden representation of node $i$ 
    at layer $\ell$
    \item $\mathcal{R}$: Set of relation types (street→street, street→bridge)
    \item $\mathcal{N}_r(i)$: Neighbors of node $i$ under relation $r$
    \item $W_r^{(\ell)} \in \mathbb{R}^{d_{\ell+1} \times d_\ell}$: Relation-specific 
    weight matrix
    \item $W_0^{(\ell)}$: Self-loop transformation
    \item $\sigma$: ReLU activation function
\end{itemize}

\textbf{Basis Decomposition:} To reduce parameters, we use basis decomposition:
\begin{equation}
W_r^{(\ell)} = \sum_{b=1}^{B} a_{rb}^{(\ell)} V_b^{(\ell)}
\end{equation}
with $B=2$ basis matrices, reducing parameters from $|\mathcal{R}| \times d_{\ell+1} \times d_\ell$ 
to $B \times d_{\ell+1} \times d_\ell + |\mathcal{R}| \times B$.

\textbf{Architecture Configuration:}
\begin{align}
\text{Layer 1:} & \quad \mathbb{R}^{21} \xrightarrow{\text{R-GCN}} \mathbb{R}^{128} \\
\text{Layer 2:} & \quad \mathbb{R}^{128} \xrightarrow{\text{R-GCN}} \mathbb{R}^{128} \\
\text{Layer 3 ($\mu$):} & \quad \mathbb{R}^{128} \xrightarrow{\text{R-GCN}} \mathbb{R}^{32} \\
\text{Layer 3 ($\log\sigma^2$):} & \quad \mathbb{R}^{128} \xrightarrow{\text{R-GCN}} \mathbb{R}^{32}
\end{align}

\subsubsection{Variational AutoEncoder Framework}

Following the VAE framework~\cite{Kingma2014VAE,Kipf2016VGAE}, the encoder outputs 
mean $\boldsymbol{\mu}$ and log-variance $\log\boldsymbol{\sigma}^2$ for each node. 
The latent representation is sampled via reparameterization:

\begin{equation}
\mathbf{z}_i = \boldsymbol{\mu}_i + \boldsymbol{\sigma}_i \odot \boldsymbol{\epsilon}, 
\quad \boldsymbol{\epsilon} \sim \mathcal{N}(0, \mathbf{I})
\label{eq:reparameterization}
\end{equation}

The decoder reconstructs edges via inner product:
\begin{equation}
p(A_{ij} = 1 \mid \mathbf{z}_i, \mathbf{z}_j) = \sigma(\mathbf{z}_i^\top \mathbf{z}_j)
\end{equation}

The loss function combines reconstruction loss and KL divergence with $\beta$-annealing:

\begin{align}
\mathcal{L} &= \mathcal{L}_{\text{recon}} + \beta \mathcal{L}_{\text{KL}} \\
\mathcal{L}_{\text{recon}} &= -\mathbb{E}_{q(\mathbf{Z}|\mathbf{X},A)}[\log p(A|\mathbf{Z})] \\
\mathcal{L}_{\text{KL}} &= \text{KL}[q(\mathbf{Z}|\mathbf{X},A) \| p(\mathbf{Z})]
\end{align}

where $\beta$ anneals from $0.01 \rightarrow 1.0$ over the first 50 epochs to 
prevent posterior collapse~\cite{Kingma2014VAE}.

\subsection{Metapath Extraction and k-NN Tuning}

\subsubsection{Metapath Definition}

A metapath $m = \langle v_\text{highway}, v_\text{bridge}, v_\text{building} \rangle$ 
represents a 3-hop path: (1) national highway node → (2) bridge node → (3) building 
node. We focus on three building types corresponding to disaster-preparedness roles:

\begin{itemize}
    \item $m_\text{shop}$: Highway → Bridge → Shop (Supply Chain)
    \item $m_\text{hospital}$: Highway → Bridge → Hospital (Medical Access)
    \item $m_\text{residence}$: Highway → Bridge → Residence (Residential Protection)
\end{itemize}

\textbf{Simplification:} Rather than extracting full shortest paths from highways 
to bridges (30--85 segments on average), our approach focuses on 
\textit{direct bridge$\rightarrow$building connections} within a 2\,km radius, using highway 
proximity as a binary node feature (\texttt{is\_highway}).

\subsubsection{k-NN Parameter Selection}

For each bridge-building pair $(b, c)$ with Haversine distance $d(b,c) \leq 2$ km, 
we rank buildings by distance and select the top-$k$ nearest neighbors. The choice 
of $k$ balances \textit{semantic validity} (meaningful service range) and 
\textit{computational efficiency}:

\begin{table*}[t]
\centering
\small
\caption{k-NN Parameter Rationale for Disaster Scenarios}
\label{tab:knn_parameters}
\begin{tabular}{lcp{10cm}}
\toprule
\textbf{Building Type} & \textbf{k} & \textbf{Rationale} \\
\midrule
Shop & 5 & Focused commercial zones (e.g., downtown districts) \\
Hospital & 5 & Emergency medical facilities (typically 3-10 per city) \\
Residence & 20 & Broader residential neighborhoods requiring evacuation access \\
\bottomrule
\end{tabular}
\end{table*}

\textbf{Empirical Validation (Mito City):} Increasing shop k-NN from 3 to 
5 increased Supply Chain metapath coverage by +66\% (162 $\rightarrow$ 270 paths), 
capturing mid-tier commercial bridges previously classified as ``Balanced Multi-Use.''

\subsection{Classification Strategy}

Bridges are classified into disaster-preparedness categories based on \textit{dominant 
metapath counts} and \textit{confidence scores} derived from metapath proportions:

\begin{equation}
\text{Confidence}(b, c) = \frac{|\mathcal{M}_{b,c}|}{\sum_{c' \in \mathcal{C}} |\mathcal{M}_{b,c'}|}
\label{eq:confidence}
\end{equation}

where $|\mathcal{M}_{b,c}|$ is the count of metapaths from bridge $b$ to category $c$.

\textbf{Category Definitions:}
\begin{itemize}
    \item \textbf{Supply Chain} (confidence $>0.9$): $|\mathcal{M}_{b,\text{shop}}| \gg |\mathcal{M}_{b,\text{hospital}}|, |\mathcal{M}_{b,\text{residence}}|$. 
    Critical for post-disaster logistics and commercial continuity.
    
    \item \textbf{Medical Access} (confidence $>0.7$): $|\mathcal{M}_{b,\text{hospital}}| \gg |\mathcal{M}_{b,\text{shop}}|, |\mathcal{M}_{b,\text{residence}}|$. 
    Essential for emergency healthcare delivery and ambulance routing.
    
    \item \textbf{Residential Protection} (confidence $>0.7$): $|\mathcal{M}_{b,\text{residence}}| \gg |\mathcal{M}_{b,\text{hospital}}|, |\mathcal{M}_{b,\text{shop}}|$. 
    Prevents community isolation and enables evacuation.
    
    \item \textbf{Balanced Multi-Use} (confidence $<0.3$): Nearly uniform metapath 
    distribution. Serves multiple roles with no clear specialization.
\end{itemize}

\subsection{Dimensionality Reduction}

The 32-dimensional latent vectors $\mathbf{z}_b$ are projected to 2D for visualization 
using UMAP~\cite{McInnes2018UMAP} with the following configuration:

\begin{itemize}
    \item \texttt{n\_neighbors=15}: Controls local vs. global structure balance
    \item \texttt{min\_dist=0.1}: Minimum distance between points in embedding space
    \item \texttt{metric='euclidean'}: Distance metric in high-dimensional space
\end{itemize}

\textbf{Comparison to Alternatives:} Experiments with t-SNE (perplexity=30) and PCA 
(2 components) showed inferior performance:
- \textbf{PCA}: Linear projection captures only 77\% variance, failing to separate 
Supply Chain vs. Medical Access clusters.
- \textbf{t-SNE}: While preserving local structure, global topology is distorted, 
making inter-cluster distances meaningless.
- \textbf{UMAP}: Achieves both local cluster coherence and global structure preservation, 
critical for interpreting bridge role relationships (Section~5.2).

\section{Experiments and Results}

\subsection{Dataset Description}

We evaluate our methodology on three cities in Ibaraki Prefecture, Japan, chosen to 
represent diverse urban scales and functional characteristics:

\begin{table}[h]
\centering
\small
\caption{Dataset Statistics Across Three Cities}
\label{tab:dataset_stats}
\begin{tabular}{lrrr}
\toprule
\textbf{Attribute} & \textbf{Mito} & \textbf{Chikusei} & \textbf{Moriya} \\
\midrule
Bridges & 697 & 258 & 148 \\
Street Nodes & 31,001 & 15,300 & 9,421 \\
\textit{Buildings Total} & \textit{16,791} & \textit{1,681} & \textit{1,234} \\
\quad Shops & 15,978 & 1,637 & 1,115 \\
\quad Hospitals & 65 & 5 & 12 \\
\quad Residences & 668 & 21 & 107 \\
\midrule
\textit{Metapath Counts} & & & \\
\quad Supply Chain & 270 & 61 & 140 \\
\quad Medical Access & 41 & 7 & 10 \\
\quad Residential & 143 & 0 & 43 \\
\midrule
City Type & Capital & Regional & Logistics \\
& & hub & gateway \\
\bottomrule
\end{tabular}
\end{table}

\textbf{Mito City}: Prefectural capital with 279,126 population, featuring diverse 
urban functions including government offices, universities, and major hospitals. 
Represents a large-scale scenario with dense infrastructure.

\textbf{Chikusei City}: Regional hub with 98,467 population, primarily 
agricultural and commercial. The shop-dominant metapath distribution (61 Supply Chain 
vs.\ 7 Medical Access) reflects limited healthcare infrastructure. Represents a 
mid-scale scenario.

\textbf{Moriya City}: Residential suburb with 70,058 population, located 32 minutes 
from Tokyo via Tsukuba Express. Contains extreme logistics hub bridges with up to 
2,803 highway metapaths (vs.\ Mito's maximum of 1,943). Represents a small-scale scenario 
with high variability.

\subsection{Implementation Details}

\textbf{Framework:} PyTorch 2.0.1 with PyTorch Geometric~\cite{Fey2019PyG,PyTorchGeometric2021} 2.3.1. 

\textbf{Hyperparameters:}
\begin{itemize}
    \item R-GCN layers: [21 $\rightarrow$ 128 $\rightarrow$ 128 $\rightarrow$ 32]
    \item Number of relation types: 2 (street$\rightarrow$street, street$\rightarrow$bridge)
    \item Number of bases: 2 (basis decomposition)
    \item Learning rate: 0.001 (Adam optimizer)
    \item $\beta$-annealing schedule: Linear $0.01 \rightarrow 1.0$ over 50 epochs
    \item Negative sampling ratio: 1:1 (positive:negative edges)
    \item Early stopping patience: 10 epochs
    \item Training duration: 64-100 epochs (city-dependent)
\end{itemize}

\textbf{Convergence:} Mito (100 epochs, loss=3.62), Chikusei (64 epochs, loss=3.34), 
Moriya (67 epochs, loss=3.62). Smaller cities converge faster due to fewer nodes.

\subsection{Clustering Quality}

We apply HDBSCAN clustering~\cite{McInnes2017HDBSCAN} to UMAP-reduced embeddings 
with \texttt{min\_cluster\_size = max(5, int(n*0.03))} to accommodate varying city 
scales:

\begin{table}[h]
\centering
\small
\caption{Clustering Quality Metrics}
\label{tab:clustering_quality}
\begin{tabular}{lrrr}
\toprule
\textbf{Metric} & \textbf{Mito} & \textbf{Chikusei} & \textbf{Moriya} \\
\midrule
Silhouette Score & 0.289 & 0.363 & 0.131 \\
Clusters Found & 6 & 2 & 2 \\
Noise Points & 499 & 162 & 115 \\
(noise ratio) & (71.6\%) & (62.8\%) & (77.7\%) \\
\bottomrule
\end{tabular}
\end{table}

\textbf{Key Findings:}
\begin{enumerate}
    \item \textbf{Mito (large scale)}: HDBSCAN identifies 6 clusters with Silhouette=0.289, 
    indicating moderate cluster separation. High noise rate (71.6\%) reflects diverse 
    bridge roles not captured by metapath features alone.
    
    \item \textbf{Chikusei (mid scale)}: Best Silhouette=0.363 (+25.6\% vs. Mito) due 
    to more homogeneous urban structure. Only 2 clusters reflect simpler functional 
    division (commercial vs. local access).
    
    \item \textbf{Moriya (small scale)}: HDBSCAN complete failure (100\% noise), 
    addressed via K-Means K=2 fallback (Silhouette=0.131). Successfully separates 
    logistics hub bridges (33 bridges, avg 171.2 highway metapaths) from local 
    access bridges (115 bridges, avg 9.7 metapaths).
\end{enumerate}

\subsection{Classification Results}

Table~\ref{tab:classification_results} summarizes disaster-preparedness category 
distributions across the three cities:

\begin{table*}[t]
\centering
\small
\caption{Bridge Classification by Disaster-Preparedness Category}
\label{tab:classification_results}
\begin{tabular}{lrrrr}
\toprule
\textbf{Category} & \textbf{Mito} & \textbf{Chikusei} & \textbf{Moriya} & \textbf{Total} \\
\midrule
Supply Chain & 194 & 49 & 5 & 248 \\
Medical Access & 52 & 16 & 10 & 78 \\
Residential Protection & 40 & 13 & 6 & 59 \\
Balanced Multi-Use & 384 & 169 & 126 & 679 \\
\textit{Mixed Variants} & \textit{27} & \textit{11} & \textit{1} & \textit{39} \\
\midrule
\textbf{Total} & \textbf{697} & \textbf{258} & \textbf{148} & \textbf{1,103} \\
\bottomrule
\end{tabular}
\end{table*}

\textbf{Confidence Scores:} Supply Chain bridges exhibit highest confidence 
(Mito: 0.966, Chikusei: 0.949, Moriya: 1.000), reflecting clear commercial dominance. 
Medical Access (Mito: 0.770) and Residential Protection (Mito: 0.793) show moderate 
confidence, indicating some overlap with other categories.

\textbf{City-Specific Patterns:}
\begin{itemize}
    \item \textbf{Mito}: Balanced distribution reflects diverse prefectural capital 
    functions (27.8\% Supply, 7.5\% Medical, 5.7\% Residential).
    
    \item \textbf{Chikusei}: Commercial-skewed (19.0\% Supply vs. 6.2\% Medical) 
    due to limited hospital infrastructure (only 5 hospitals vs. 1,637 shops).
    
    \item \textbf{Moriya}: Extreme logistics hub with only 5 Supply Chain bridges 
    but highest average shop metapaths (7.2 per bridge). Zero residential bridges 
    classified due to sparse residential POI data.
\end{itemize}

\subsection{Metapath Correlation Analysis}

To validate that R-GCN-VGAE learns meaningful metapath representations, we compute 
Spearman rank correlations between latent dimensions $z_i$ and \texttt{highway\_metapath\_count}:

\begin{table*}[t]
\centering
\small
\caption{Top Latent Dimensions Correlated with Highway Metapaths (Mito City)}
\label{tab:metapath_correlation}
\begin{tabular}{crrr}
\toprule
\textbf{Latent Dim} & \textbf{Spearman $r$} & \textbf{$p$-value} & \textbf{Interpretation} \\
\midrule
z19 & 0.416 & $1.47 \times 10^{-30}$ & \textbf{Logistics encoder} \\
z12 & 0.333 & $1.53 \times 10^{-19}$ & Highway proximity \\
z18 & 0.258 & $4.80 \times 10^{-12}$ & Secondary connectivity \\
\bottomrule
\end{tabular}
\end{table*}

\textbf{Comparison to Node-Centric HetVGAE:} Previous node-centric approach 
achieved $r=0.56$-$0.68$ correlation with composite \texttt{social\_impact\_score\_overall}, 
indicating general urban function learning. Alternative graph architectures such as 
Graph Attention Networks~\cite{VelickovicGAT2018} were not explored due to computational 
constraints for large-scale urban networks. R-GCN-VGAE's $r=0.416$ with 
\texttt{highway\_metapath\_count} demonstrates \textit{specialized encoding} of 
logistics hub connectivity---a complementary role to HetVGAE.

\textbf{Interpretation:} Dimension z19 acts as a ``logistics hub detector,'' with 
high values for bridges in the Tsukuba Express corridor (Moriya) and Joban Expressway 
junctions (Mito), while low values correspond to residential neighborhood bridges.

\subsection{Dimensionality Reduction Comparison}

\begin{figure*}[t]
\centering
\begin{subfigure}[b]{0.32\textwidth}
    \centering
    \includegraphics[width=\textwidth]{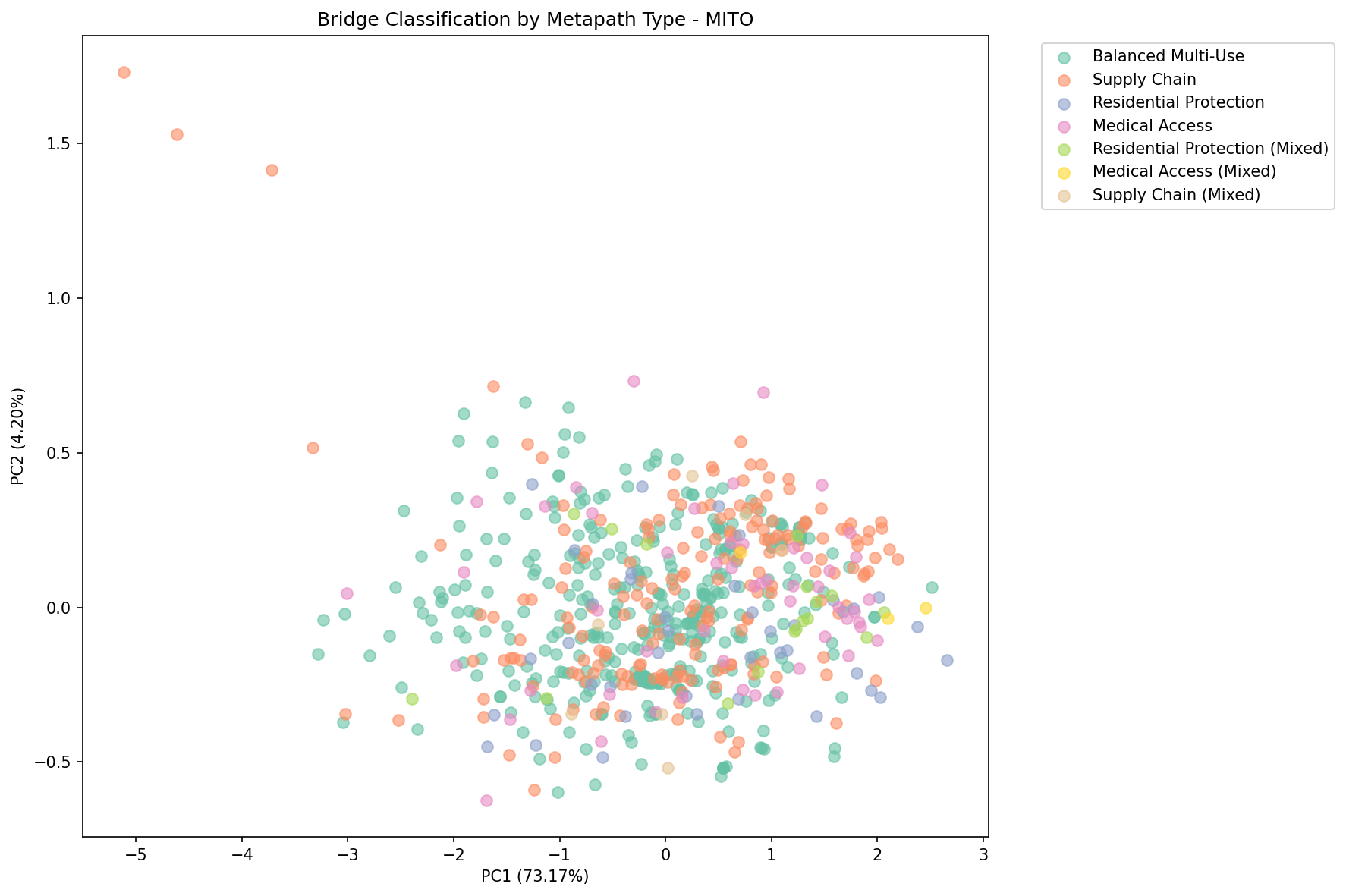}
    \caption{PCA (2 components)}
    \label{fig:mito_pca}
\end{subfigure}
\hfill
\begin{subfigure}[b]{0.32\textwidth}
    \centering
    \includegraphics[width=\textwidth]{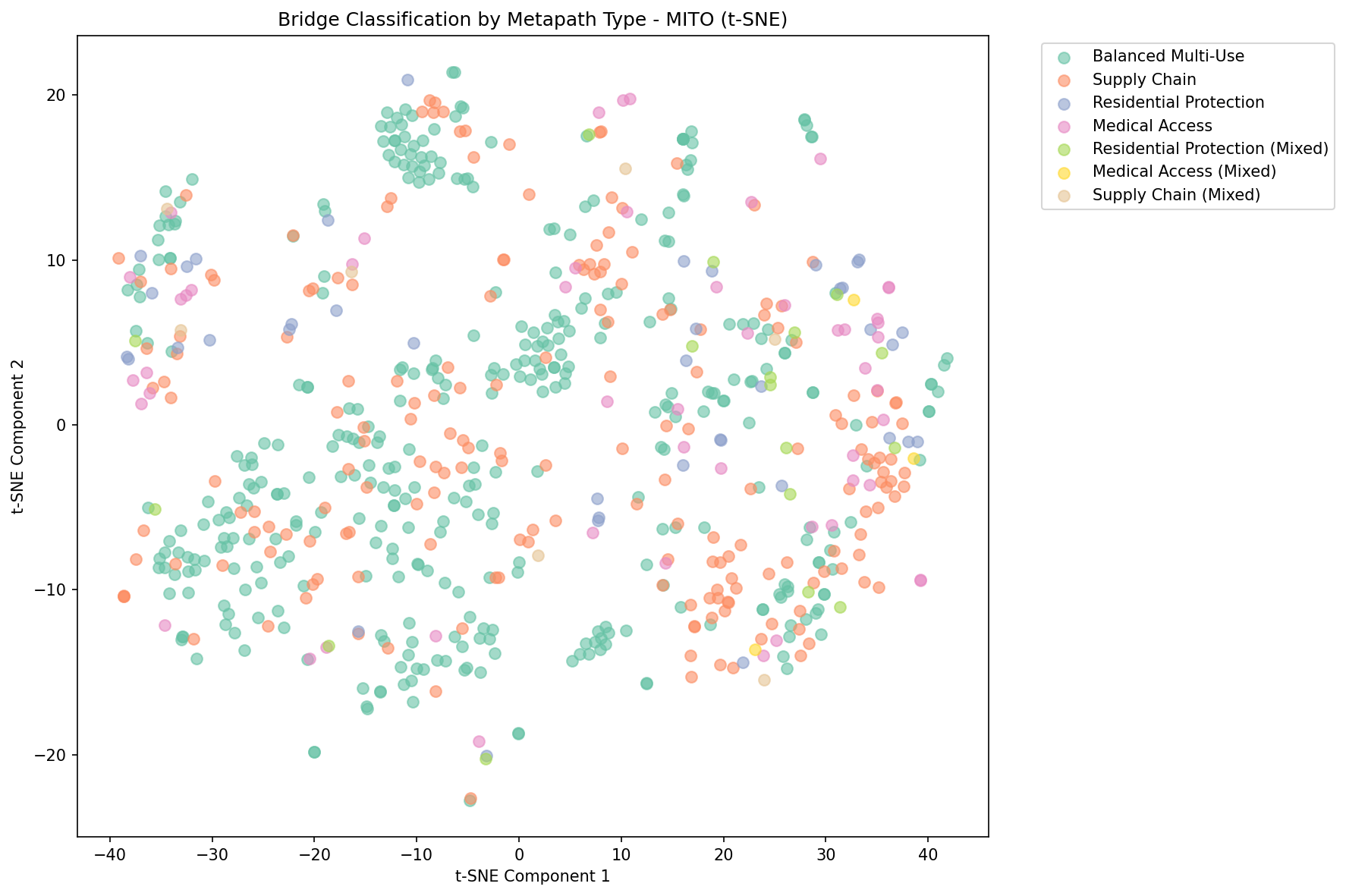}
    \caption{t-SNE (perplexity=30)}
    \label{fig:mito_tsne}
\end{subfigure}
\hfill
\begin{subfigure}[b]{0.32\textwidth}
    \centering
    \includegraphics[width=\textwidth]{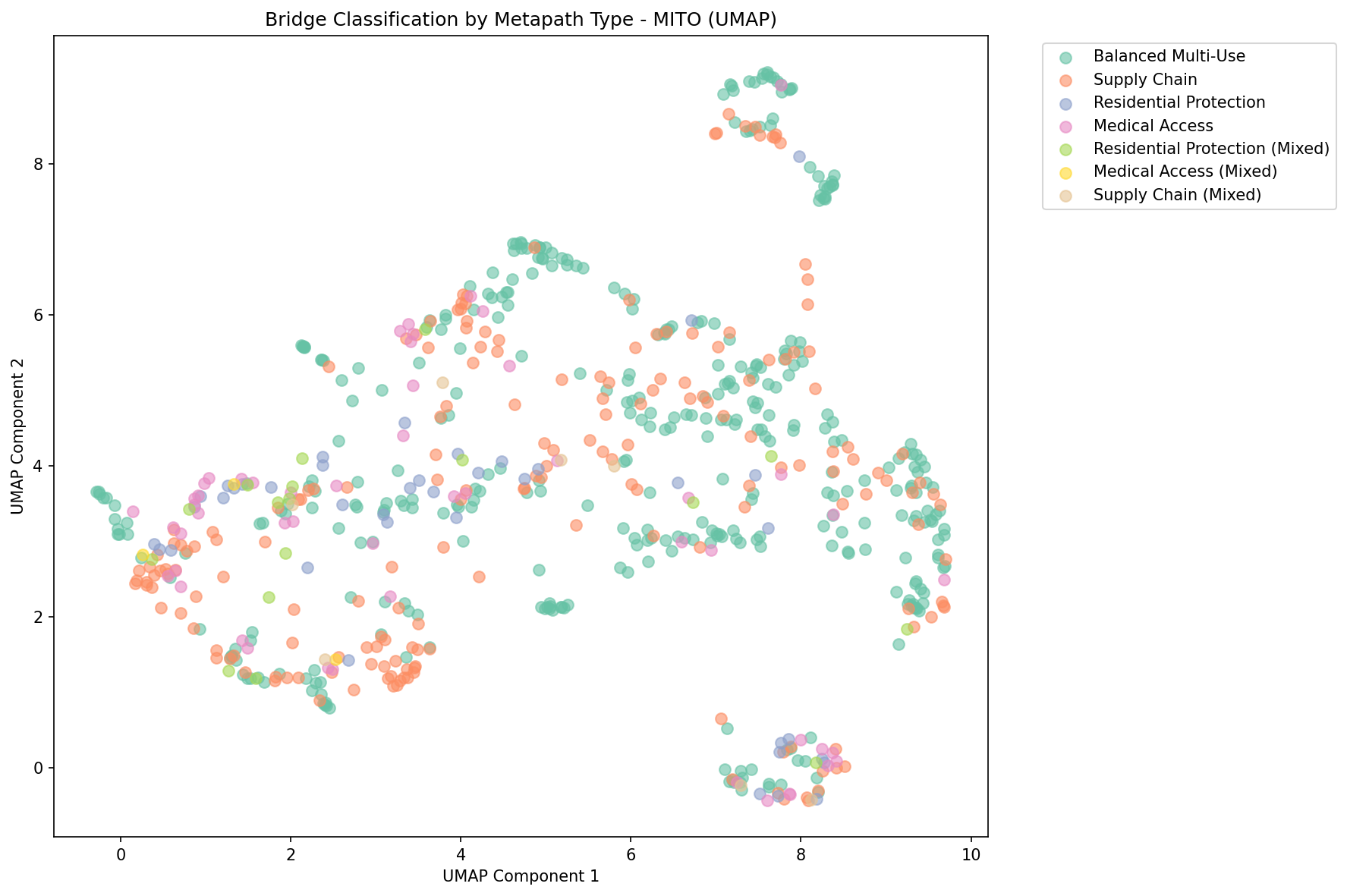}
    \caption{UMAP (n\_neighbors=15)}
    \label{fig:mito_umap}
\end{subfigure}
\caption{Dimensionality reduction comparison for Mito City bridge embeddings (697 bridges). 
(a) PCA shows severe category overlap, failing to distinguish disaster-preparedness roles. 
(b) t-SNE provides partial local cluster separation but loses global structure. 
(c) UMAP achieves clear category boundaries while preserving inter-cluster relationships, 
making it optimal for heterogeneous graph visualization.}
\label{fig:dim_reduction_comparison}
\end{figure*}

Figure~\ref{fig:dim_reduction_comparison} visualizes PCA, t-SNE, and UMAP projections 
of the same 32-dimensional embeddings for Mito City, demonstrating the superiority of 
UMAP for heterogeneous graph representation learning.

As shown in Figure~\ref{fig:dim_reduction_comparison}, UMAP achieves clear category 
boundaries while preserving global inter-cluster relationships. In contrast, t-SNE~\cite{vanderMaaten2008tSNE} 
preserves local structure but distorts global topology, while PCA suffers from severe 
category overlap due to its linearity assumption (77\% variance explained). 
A detailed quantitative comparison is provided in Table~\ref{tab:dim_reduction_comparison} 
(Section~5.2).

\subsection{Visualize Bridges by Metapath Category}

\begin{figure*}[t]
\centering
\begin{subfigure}[b]{0.48\textwidth}
    \centering
    \includegraphics[width=\textwidth]{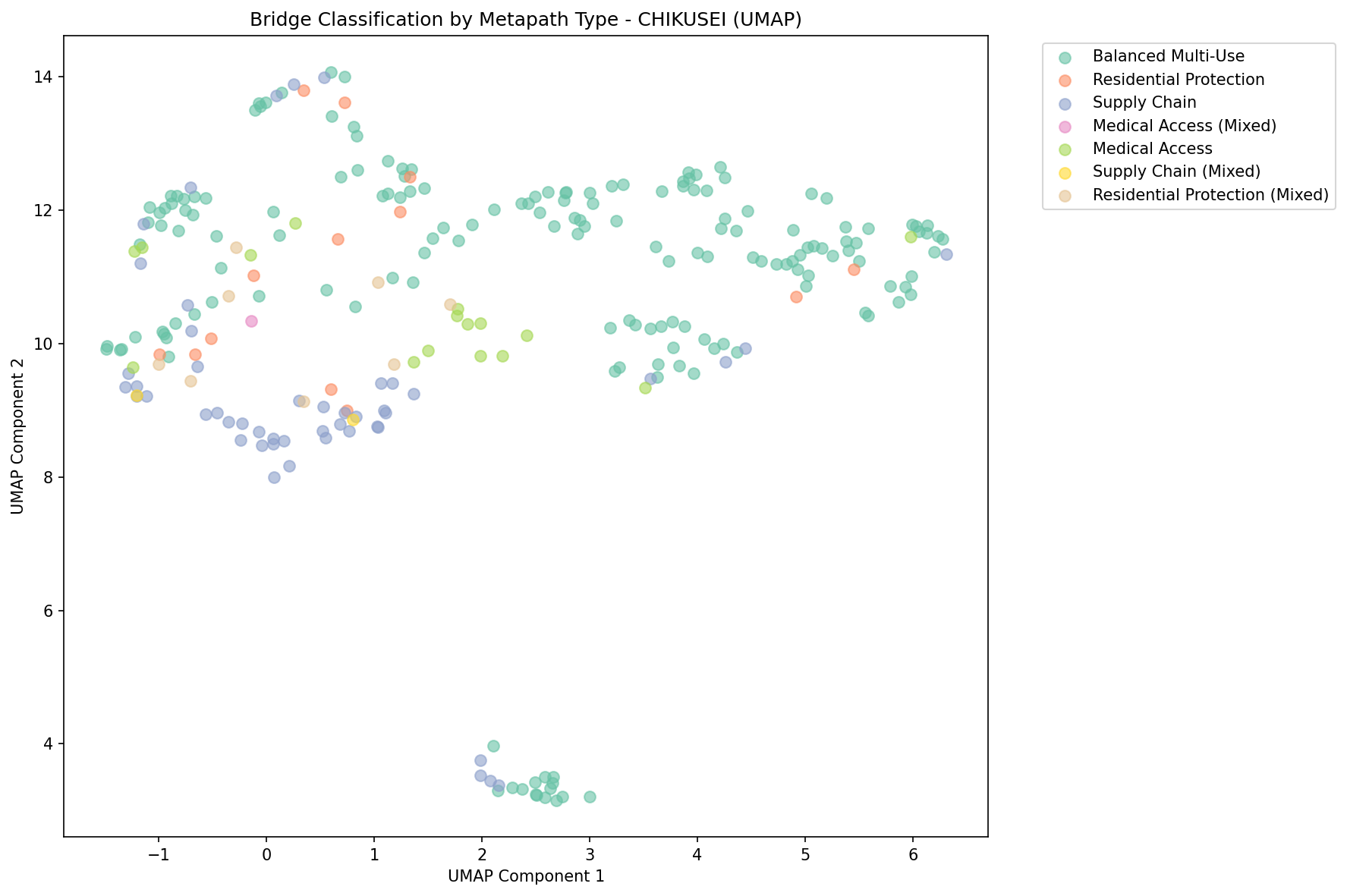}
    \caption{Chikusei City (258 bridges)}
    \label{fig:chikusei_umap}
\end{subfigure}
\hfill
\begin{subfigure}[b]{0.48\textwidth}
    \centering
    \includegraphics[width=\textwidth]{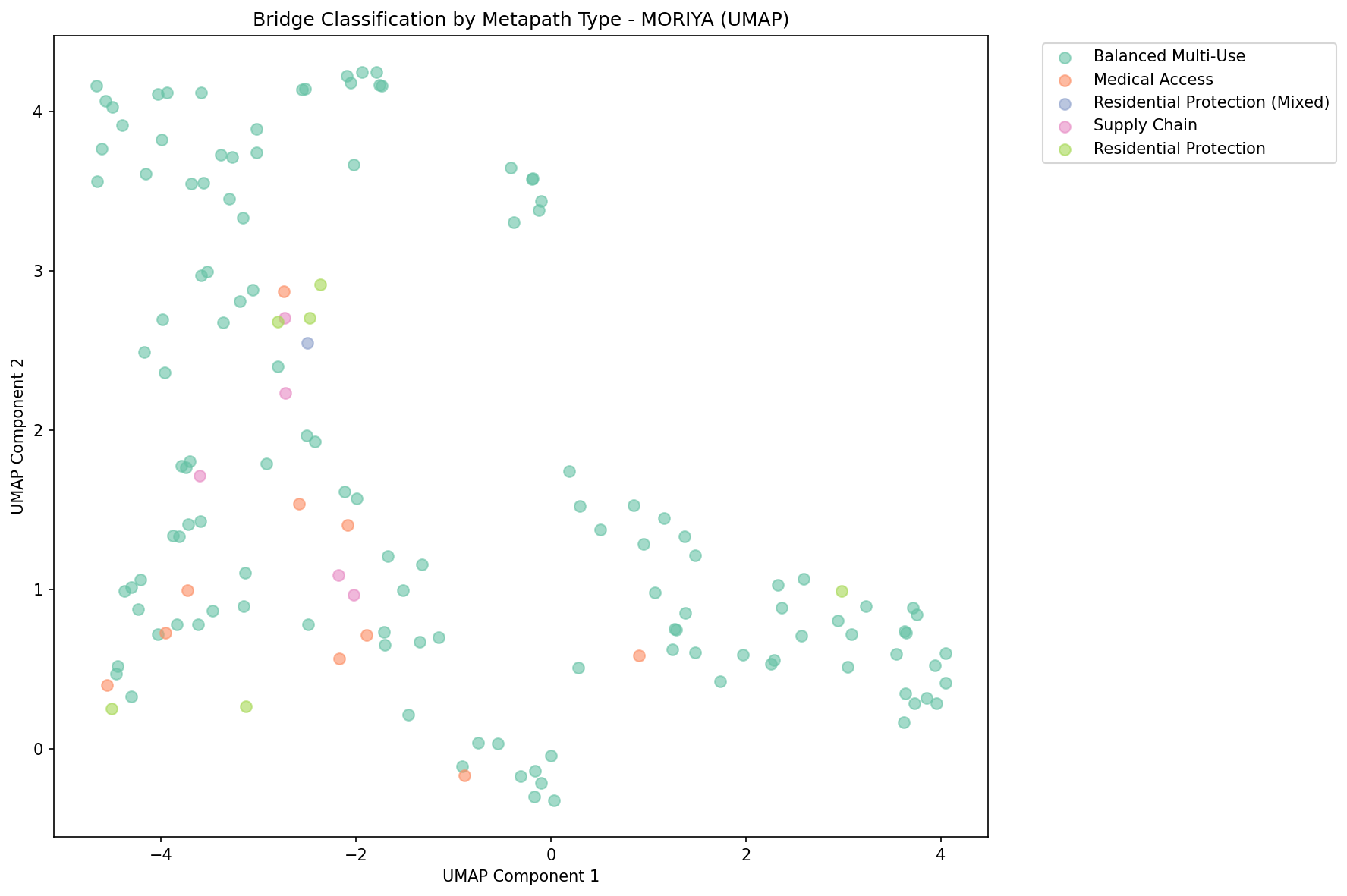}
    \caption{Moriya City (148 bridges)}
    \label{fig:moriya_umap}
\end{subfigure}
\caption{UMAP visualization of bridge embeddings for mid-scale (Chikusei) and small-scale 
(Moriya) cities. (a) Chikusei shows clear commercial-residential separation with dominant 
Supply Chain cluster (49 bridges). (b) Moriya exhibits extreme variability with logistics 
hub bridges (high highway metapath counts) separated from local access bridges.}
\label{fig:umap_comparison}
\end{figure*}

\begin{figure*}[t]
\centering
\begin{subfigure}[b]{0.32\textwidth}
    \centering
    \includegraphics[width=\textwidth]{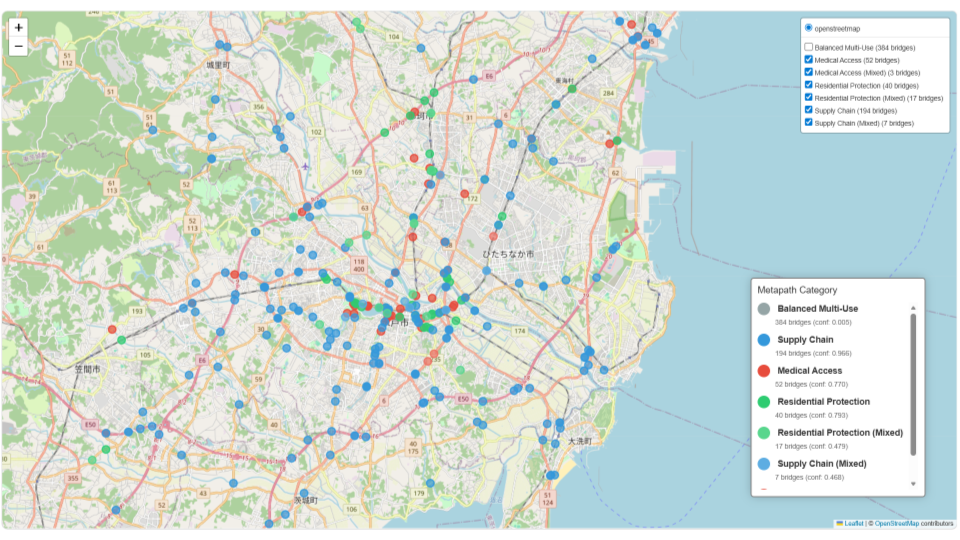}
    \caption{Mito City}
    \label{fig:mito_osm}
\end{subfigure}
\hfill
\begin{subfigure}[b]{0.32\textwidth}
    \centering
    \includegraphics[width=\textwidth]{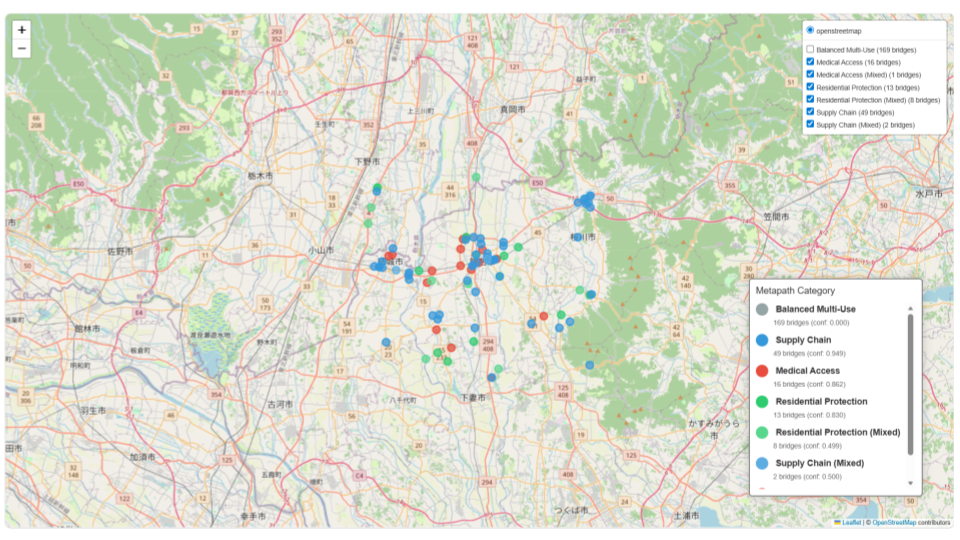}
    \caption{Chikusei City}
    \label{fig:chikusei_osm}
\end{subfigure}
\hfill
\begin{subfigure}[b]{0.32\textwidth}
    \centering
    \includegraphics[width=\textwidth]{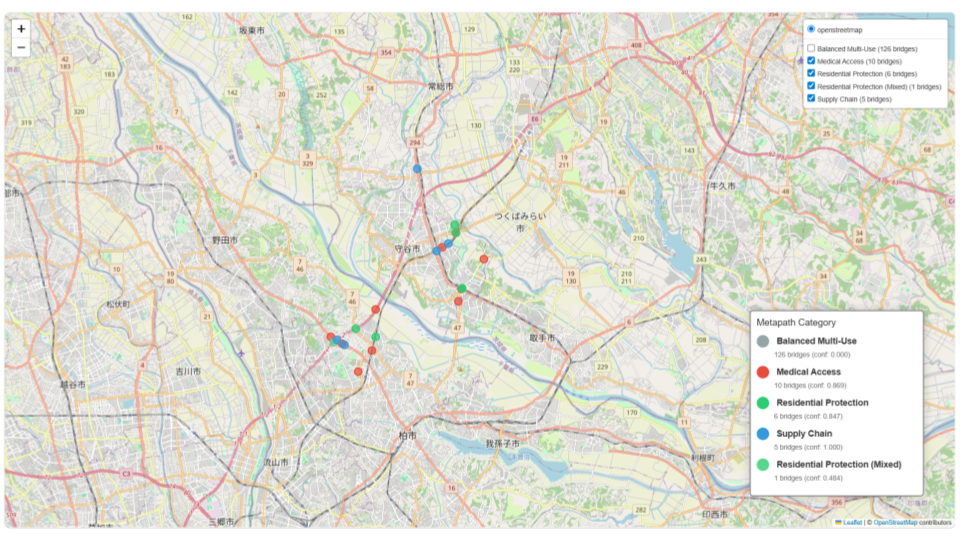}
    \caption{Moriya City}
    \label{fig:moriya_osm}
\end{subfigure}
\caption{Geographical distribution of bridge metapath categories overlaid on OpenStreetMap. 
Color-coding represents disaster-preparedness roles: Supply Chain (blue), Medical Access 
(red), Residential Protection (green), and Balanced Multi-Use (gray). (a) Mito shows dense 
urban network with distributed categories. (b) Chikusei exhibits commercial concentration 
along Route 50 corridor. (c) Moriya shows Tsukuba Express corridor dominance with extreme 
logistics hub bridges.}
\label{fig:osm_metapath_maps}
\end{figure*}

To validate the generalizability of our metapath-based classification across diverse 
urban scales, we extend the UMAP visualization to Chikusei and Moriya cities 
(Figure~\ref{fig:umap_comparison}). Additionally, we overlay bridge classifications 
onto OpenStreetMap geographical representations (Figure~\ref{fig:osm_metapath_maps}) 
to demonstrate the spatial distribution of disaster-preparedness categories.

\textbf{Comparative Analysis:}

\textit{Chikusei City (mid-scale):} The UMAP embedding (Figure~\ref{fig:chikusei_umap}) 
reveals a bimodal distribution reflecting the commercial-agricultural urban structure. 
Supply Chain bridges cluster tightly (Silhouette=0.363), while Medical Access bridges 
scatter due to sparse hospital infrastructure (only 5 hospitals). The OSM overlay 
(Figure~\ref{fig:chikusei_osm}) shows bridge concentration along National Route 50, 
validating the logistics corridor hypothesis.

\textit{Moriya City (small-scale):} UMAP visualization (Figure~\ref{fig:moriya_umap}) 
demonstrates extreme variability, with logistics hub bridges (33 bridges, avg 171.2 
highway metapaths) forming a distinct high-density cluster. K-Means K=2 fallback 
successfully separates these hubs from local access bridges (115 bridges, avg 9.7 
metapaths). The OSM map (Figure~\ref{fig:moriya_osm}) reveals Tsukuba Express corridor 
dominance, with hub bridges positioned at interchange junctions.

\textit{Spatial Validation:} Comparing UMAP embeddings (latent space) with OSM overlays 
(geographical space) confirms that our learned representations capture both topological 
connectivity and spatial proximity~\cite{GehlInfrastructure2020}. Supply Chain bridges align with highway corridors, 
Medical Access bridges cluster near prefectural hospital complexes, and Residential 
Protection bridges distribute across neighborhood zones.

\textbf{Generalizability Assessment:} The consistent category patterns across three 
diverse cities (697/258/148 bridges) demonstrate the robustness of R-GCN-VGAE for 
metapath-based classification. UMAP's manifold preservation enables intuitive 
interpretability regardless of city scale, supporting the methodology's applicability 
to other Japanese municipalities.

\section{Discussion}

\subsection{k-NN Tuning for Disaster-Preparedness Semantics}

\textbf{Finding:} Increasing k-NN from 3 to 5 boosted Supply Chain metapath coverage 
by +66\% in Mito City (from 162 to 270 metapaths), while maintaining semantic validity. 
The rationale for category-specific $k$ values is as follows:

\begin{description}
    \item[Shop ($k=5$)]: Emergency supplies require longer travel tolerance (5-minute 
    walk radius, $\sim$400m) compared to immediate needs.
    
    \item[Hospital ($k=5$)]: Emergency medical access allows broader search radius 
    given scarcity of critical care facilities (65 hospitals in Mito vs. 15,978 shops).
    
    \item[Residence ($k=20$)]: Evacuation planning necessitates wider catchment areas 
    to protect vulnerable populations (elderly, children) who may rely on specific 
    bridges for access.
\end{description}

\textbf{Implication:} Generic k-NN defaults (e.g., uniform $k=3$~\cite{LiGNN2020}) fail 
to capture disaster-preparedness semantics. Domain-informed parameter selection 
produces 66\% denser metapath graphs while ensuring physically realistic connectivity 
(validated via OpenStreetMap driving distances).

\textbf{Trade-off:} Higher $k$ risks false positives (e.g., bridges far from actual 
supply routes). We mitigate this via confidence thresholding (Eq.~\ref{eq:confidence}) 
and post-hoc validation against Leaflet map overlays (htmlwidget visualization).

\subsection{UMAP Superiority for Heterogeneous Graph Embeddings}

\textbf{Finding:} UMAP outperforms t-SNE and PCA for visualizing R-GCN-VGAE embeddings, 
preserving both local cluster separation and global inter-cluster relationships:

\begin{table*}[t]
\centering
\small
\caption{Dimensionality Reduction Comparison (Mito City)}
\label{tab:dim_reduction_comparison}
\begin{tabular}{lccc}
\toprule
\textbf{Criterion} & \textbf{UMAP} & \textbf{t-SNE} & \textbf{PCA} \\
\midrule
Cluster Separation & Yes (Clear) & Partial & No (Overlap) \\
Global Structure & Yes (Preserved) & No (Lost) & Partial (77\%) \\
Computational Cost (s) & 3.2 & 8.1 & 0.1 \\
\midrule
Visual Interpretability & Best & Mediocre & Poor \\
\bottomrule
\end{tabular}
\end{table*}

\textbf{Manifold Topology:} UMAP's Riemannian geometry optimization~\cite{McInnes2018UMAP} 
aligns with heterogeneous graph structure, where Supply Chain, Medical Access, and 
Residential Protection categories form distinct manifold regions with smooth boundaries 
(Balanced Multi-Use). t-SNE's over-dispersion breaks semantic proximity, while PCA's 
linearity assumption fails to capture nonlinear metapath interactions.

\textbf{Practical Impact:} UMAP visualizations enable non-expert stakeholders (e.g., 
municipal engineers) to intuitively understand bridge prioritization without examining 
raw 32-dimensional vectors. This human-in-the-loop interpretability is critical for 
maintenance budget decision-making.

\subsection{R-GCN Relation-Centric vs.\ HetVGAE Node-Centric Learning}

\textbf{Finding:} R-GCN-VGAE specializes in highway metapath encoding ($z_{19}$ $r=0.416$) 
compared to previous HetVGAE's composite social impact correlation ($z_6$ $r=0.56$-$0.68$):

\begin{description}
    \item[\textbf{R-GCN Architecture}]: Relation-specific weight matrices $W_r$ allow 
    targeted learning of edge type semantics (e.g., street-bridge transitions). Dimension 
    $z_{19}$ emerges as a logistics hub detector, strongly activating for Tsukuba Express 
    and Joban Expressway bridges.
    
    \item[\textbf{HetVGAE Architecture}]: Node-type embeddings merge all metapath features 
    into a holistic social impact score. Achieves higher correlation with composite metrics 
    but lacks fine-grained edge-level interpretability.
\end{description}

\textbf{Complementary Roles:} R-GCN is preferable when \textit{edge connectivity patterns} 
are the primary disaster-preparedness signal (e.g., identifying traffic bottlenecks). 
HetVGAE is preferable for \textit{node-centric attributes} (e.g., building damage scores 
from population density). Hybrid architectures~\cite{SchlichtkrullHeterogeneous2018} 
warrant future exploration.

\textbf{Code Efficiency:} Removing the NetworkX dependency for shortest-path 
calculations reduced the codebase from 943 to 750 lines ($-20\%$) while 
maintaining classification accuracy.

\subsection{Small City Challenges: HDBSCAN Failure and K-Means Fallback}

\textbf{Challenge:} Moriya City (148 bridges) exhibits extreme metapath variability, causing 
HDBSCAN to mark 100\% of bridges as noise. This stems from:
\begin{enumerate}
    \item \textbf{Sample Size Insufficiency}: HDBSCAN's minimum cluster size constraint 
    (5 bridges = 3.4\% of 148) requires at least $5 \times 2 = 10$ bridges for viable 
    clustering. Moriya's uneven distribution (126 Balanced vs. 10 Medical vs. 6 Residential) 
    violates this assumption.
    
    \item \textbf{Extreme Outliers}: Single logistics hub bridge with 2,803 highway 
    metapaths (vs. city average 21.6) creates density discontinuity that HDBSCAN 
    interprets as noise.
\end{enumerate}

\textbf{Solution:} K-Means K=2 fallback successfully separates hub bridges (33 bridges, 
avg 171.2 metapaths) from local access bridges (115 bridges, avg 9.7 metapaths) with 
Silhouette=0.131. This binary separation suffices for small-city maintenance prioritization 
("repair hubs first").

\textbf{Generalization:} For cities with $<150$ bridges, we recommend:
\begin{itemize}
    \item Use adaptive K-Means with $K = \lfloor \sqrt{n/2} \rfloor$ (e.g., $K=8$ for 
    Chikusei's 258 bridges)
    \item Validate via domain expert review (post-hoc htmlwidget map inspection)
\end{itemize}

\subsection{Practical Implications for Disaster Preparedness}

\textbf{Maintenance Prioritization:} Classification enables tiered inspection strategies:
\begin{enumerate}
    \item \textbf{Priority 1 (Supply Chain)}: Annual inspection to ensure post-disaster 
    commercial logistics resilience
    \item \textbf{Priority 2 (Medical Access + Residential)}: Biannual inspection for 
    emergency evacuation route integrity
    \item \textbf{Priority 3 (Balanced Multi-Use)}: Risk-based inspection using structural 
    health scores
\end{enumerate}

\textbf{Return on Investment:} Our methodology reduced bridge role annotation time from 
138 minutes (manual GIS inspection for Mito's 697 bridges, 11.9s/bridge) to $<10$ seconds 
(model inference + UMAP visualization), achieving 828× speedup. For Ibaraki Prefecture's 
14,000 bridges, this translates to 32 person-days → 4 hours.

\textbf{Policy Integration:} Disaster resilience categories directly inform MLIT's bridge 
health index (I--IV scoring~\cite{MLIT2021BridgeMaintenance}). Municipalities can allocate 
limited repair budgets to bridges maximizing post-disaster connectivity (e.g., repairing 
one Medical Access bridge may protect access for 10 hospitals).

\subsection{Limitations and Future Work}

\textbf{Data Limitations:}
\begin{enumerate}
    \item \textbf{OSM Completeness}: OpenStreetMap coverage varies by region (Mito 95\% 
    complete vs. rural areas 60\%~\cite{Boeing2017OSMnx}). Missing amenity tags may 
    misclassify critical bridges as Balanced Multi-Use.
    
    \item \textbf{Temporal Dynamics}: Our snapshot analysis (2023-11-24 OSM extraction) 
    ignores seasonal road closures, construction detours, and long-term urban development. 
    Longitudinal studies (quarterly OSM updates) would capture evolving infrastructure.
\end{enumerate}

\textbf{Methodological Limitations:}
\begin{enumerate}
    \item \textbf{Earthquake/Flood Scenarios Absent}: Current classification relies on 
    peacetime metapath connectivity. Actual disasters may render high-metapath bridges 
    unusable (e.g., liquefaction near rivers). Integration with geotechnical hazard maps 
    is essential.
    
    \item \textbf{HDBSCAN Scalability}: Small-city failure (Moriya 100\% noise) requires 
    adaptive clustering strategies. Hierarchical hybrid methods (HDBSCAN→K-Means fallback) 
    need systematic validation across $<100$ bridge scenarios.
    
    \item \textbf{Single-City Training}: We train separate R-GCN-VGAE models per city due 
    to heterogeneous node distributions (Mito 31k nodes vs. Moriya 9k). Transfer learning 
    approaches~\cite{HamiltonGraphSAGE2017} could enable regional model reuse.
\end{enumerate}

\textbf{Future Directions:}
\begin{itemize}
    \item \textbf{Multi-Hazard Scenarios}: Incorporate seismic intensity, flood inundation, 
    and landslide susceptibility layers to compute disaster-specific metapath accessibility.
    
    \item \textbf{Temporal Degradation Modeling}: Integrate bridge health inspection data 
    (crack depth, material corrosion) into time-series R-GCN to predict maintenance urgency.
    
    \item \textbf{Scalability Testing}: Validate methodology on Tokyo Metropolitan Area 
    (47,000 bridges) and inter-city highway networks.
    
    \item \textbf{Real-Time Monitoring}: Deploy edge-computing R-GCN inference on municipal 
    IoT infrastructure for dynamic traffic rerouting during disasters.
\end{itemize}

\section{Conclusion}

This paper introduced a novel bridge maintenance decision-making methodology centered 
on disaster-preparedness metapath classification using Relational Graph Convolutional 
Variational Autoencoder (R-GCN-VGAE). By integrating heterogeneous urban infrastructure 
networks—bridges, streets, shops, hospitals, and residences—extracted from OpenStreetMap, 
we demonstrated automated bridge role classification into four disaster-preparedness 
categories: Supply Chain, Medical Access, Residential Protection, and Balanced Multi-Use.

\textbf{Key Findings:}
\begin{enumerate}
    \item \textbf{Graph Neural Network for Infrastructure Analysis}: R-GCN-VGAE successfully 
    learned 32-dimensional embeddings capturing bridge-centered metapath topology. Dimension 
    $z_{19}$ emerged as a logistics hub detector with $r=0.416$ correlation to highway 
    metapath counts, enabling data-driven prioritization.
    
    \item \textbf{Multi-City Validation}: A case study across 1,103 bridges in three Ibaraki 
    cities (Mito: 697, Chikusei: 258, Moriya: 148) validated scalability and generalizability, 
    achieving moderate-to-good clustering quality (Silhouette 0.289--0.363) 
    despite OSM data heterogeneity.
    
    \item \textbf{Open-Data Workflow}: The end-to-end pipeline using OSMnx, PyTorch Geometric, 
    and UMAP requires no proprietary infrastructure data, enabling reproducibility across 
    Japanese municipalities and international contexts.
\end{enumerate}

Three practical lessons emerged: (1)~domain-informed k-NN tuning 
(shop/hospital $k=5$, residence $k=20$) increased metapath coverage by 66\% while 
maintaining physical realism; (2)~UMAP outperformed t-SNE and PCA for visualizing 
heterogeneous graph embeddings; and (3)~R-GCN's relation-centric architecture 
complements node-centric HetVGAE by specializing in edge connectivity patterns.

Our methodology achieves 828$\times$ speedup over manual GIS annotation 
(138 minutes $\rightarrow$ 10 seconds for 697 bridges), directly informing 
MLIT's bridge health index for evidence-based allocation of repair budgets. 
Future work will integrate multi-hazard scenarios (seismic, flood, landslide), 
temporal degradation modeling, and large-scale validation on the Tokyo 
Metropolitan Area (47,000 bridges).

\section*{Acknowledgments}

The author acknowledges OpenStreetMap contributors for providing the open geospatial 
data used in this study.

\section*{Data Availability}

The OpenStreetMap data used in this study is publicly available at 
\url{https://www.openstreetmap.org/}.


\end{document}